\documentclass[conference]{IEEEtran}
\IEEEoverridecommandlockouts
\usepackage{cite}
\usepackage{amsmath,amssymb,amsfonts}
\usepackage{float}
\usepackage{algorithmic}
\usepackage{graphicx}
\usepackage{textcomp}
\usepackage{xcolor}
\usepackage{multirow}
\usepackage{svg}
\usepackage{verbatim}
\usepackage{tablefootnote}

\def\BibTeX{{\rm B\kern-.05em{\sc i\kern-.025em b}\kern-.08em
    T\kern-.1667em\lower.7ex\hbox{E}\kern-.125emX}}

\makeatletter
\newcommand{\myheader}{Accepted for publication in 2026 IEEE Intelligent Vehicles Symposium (IV)}

\def\ps@IEEEtitlepagestyle{%
  \def\@oddhead{\footnotesize\myheader\hfill}%
  \def\@evenhead{\footnotesize\myheader\hfill}%
  \def\@oddfoot{\vbox{\mycopyrightnotice\vss\hbox to \textwidth{\hss\thepage\hss}}}
  \def\@evenfoot{}%
}

\def\ps@headings{%
  \def\@oddhead{\footnotesize\myheader\hfill}%
  \def\@evenhead{\footnotesize\myheader\hfill}%
  \def\@oddfoot{\hfill\thepage\hfill}
  \def\@evenfoot{\hfill\thepage\hfill}%
}
\makeatother

\newcommand{\mycopyrightnotice}{%
  \begin{minipage}[t]{2\columnwidth+\columnsep}\vspace{8pt}
    \scriptsize © 2026 IEEE. Personal use of this material is permitted. Permission from IEEE must be obtained for all other uses, in any current or future media, including reprinting/republishing this material for advertising or promotional purposes, creating new collective works, for resale or redistribution to servers or lists, or reuse of any copyrighted component of this work in other works.
  \end{minipage}%
}

\begin{document}

\title{RADE-Net: Robust Attention Network for Radar-Only Object Detection in Adverse Weather}

\author{\IEEEauthorblockN{Christof Leitgeb}
\IEEEauthorblockA{\textit{Infineon Technologies AG} \\
\textit{Institute of Visual Computing}\\
\textit{Graz University of Technology}\\
Graz, Austria \\
christof.leitgeb@student.tugraz.at}
\and
\IEEEauthorblockN{Thomas Puchleitner}
\IEEEauthorblockA{\textit{Infineon Technologies AG} \\
Graz, Austria \\
thomas.puchleitner \\@student.tugraz.at}
\and
\IEEEauthorblockN{Max Peter Ronecker}
\IEEEauthorblockA{\textit{Virtual Vehicle Research GmbH} \\
\textit{Institute of Visual Computing}\\
\textit{Graz University of Technology}\\
Graz, Austria \\
max.ronecker@v2c2.at}
\and
\IEEEauthorblockN{Daniel Watzenig}
\IEEEauthorblockA{\textit{Institute of Visual Computing}\\
\textit{Graz University of Technology}\\
Graz, Austria \\
daniel.watzenig@tugraz.at}
}

\maketitle

\begin{abstract}
Automotive perception systems are obligated to meet high requirements. While optical sensors such as Camera and Lidar struggle in adverse weather conditions, Radar provides a more robust perception performance, effectively penetrating fog, rain, and snow. Since full Radar tensors have large data sizes and very few datasets provide them, most Radar-based approaches work with sparse point clouds or 2D projections, which can result in information loss. Additionally, deep learning methods show potential to extract richer and more dense features from low level Radar data and therefore significantly increase the perception performance. Therefore, we propose a 3D projection method for fast-Fourier-transformed 4D Range-Azimuth-Doppler-Elevation (RADE) tensors. Our method preserves rich Doppler and Elevation features while reducing the required data size for a single frame by 91.9\% compared to a full tensor, thus achieving higher training and inference speed as well as lower model complexity. We introduce RADE-Net, a lightweight model tailored to 3D projections of the RADE tensor. The backbone enables exploitation of low‑level and high‑level cues of Radar tensors with spatial and channel-attention. The decoupled detection heads predict object center-points directly in the Range-Azimuth domain and regress rotated 3D bounding boxes from rich feature maps in the cartesian scene. We evaluate the model on scenes with multiple different road users and under various weather conditions on the large-scale K-Radar dataset and achieve a 16.7\% improvement compared to their baseline, as well as 6.5\% improvement over current Radar-only models. Additionally, we outperform several Lidar approaches in scenarios with adverse weather conditions. The code is available under https://github.com/chr-is-tof/RADE-Net.

\end{abstract}


\section{Introduction}
The development of reliable and efficient algorithms for automotive perception systems is crucial for the advancement of autonomous vehicles and robots. To ensure save navigation, these systems must be able to detect and track objects in various environments and weather conditions, while running in real-time on embedded hardware with limited power, thermal headroom and memory. Therefore, environment robustness and memory efficiency are considered practical necessities. While providing high resolution and structural information, optical sensors such as Lidar and Camera perform poorly in adverse weather conditions~\cite{kurup_dsor_2021, paek_k-radar_2022, feng_review_2025, yoneda_automated_2019}. Automotive FMCW Radar systems, which operate at larger wavelengths, can penetrate fog and rain while providing accurate depth information, radial velocity, as well as continuously improving angular resolution~\cite{feng_review_2025, yoneda_automated_2019, paek_k-radar_2022, rebut_raw_2022, lim_radical_2021, xm40-jx59-22, ouaknine_carrada_2021}.
\IEEEpubidadjcol
Several existing Object Detection approaches use Radar as a complementary sensor to enhance Camera- or Lidar-features with motion information or weather robustness~\cite{peng_mutualforce_2025, huang_l4dr_2025, fent_dpft_2024, peng_moral_2025, wolters_sparc_2025, liu_echoes_nodate}. Radar data is typically used in point cloud format which comes with small data-sizes and analogous processing techniques to Lidar data. Conventionally, Radar point clouds are generated by applying adaptive thresholding (e.g., CFAR) and clustering techniques to the 4-dimensional FFT-processed Radar tensor~\cite{lim_radical_2021}. This results in very few points per object, especially for vulnerable road users such as pedestrians and cyclists. The lack of structural information leads to poor performance for Radar-only object detection and classification~\cite{yao_exploring_2024}. In order to overcome these challenges, several approaches have used the capabilities of modern deep learning models to learn richer details from raw Radar ADC data~\cite{yang_adcnet_2023, giroux_t-fftradnet_2023, rebut_raw_2022} or FFT-processed data~\cite{gao_ramp-cnn_2021, cheng_transrad_2025, dalbah_transradar_2024, jia_tcradar_2024}. In order to avoid the large memory requirements of loading full 4D Radar Range-Azimuth-Doppler-Elevation tensors for each frame, many approaches relied on 2D projections~\cite{gao_ramp-cnn_2021, cheng_transrad_2025, dalbah_transradar_2024, jia_tcradar_2024}, sparse tensors~\cite{paek_k-radar_2022, kong_rtnh_2023, paek_enhanced_2023}, or reduced tensors~\cite{ding_radarocc_2024}. However, 2D projections lack the direct, inherent relationship between positional and velocity features present in the Radar tensor. Sparse or reduced tensor methods explore those relationships at the cost of computationally expensive model architectures~\cite{ding_radarocc_2024, kong_rtnh_2023}.

We propose a hybrid solution which preserves the direct relationships between Range-Azimuth (RA) positional bins and Doppler velocity features, as well as RA bins and Elevation features in form of 3D projections. While the Doppler spectrum contains important object-characteristic motion patterns~\cite{gusland_improving_2024}, Elevation features provide information for true vertical object discrimination, thereby improving robustness and classification performance while reducing false positives~\cite{jing_elevation_2025}. The CNN based encoder-decoder structure resembles a UNet~\cite{navab_u-net_2015} which has become very popular when processing "irregular shaped" inputs such as MRI/CT images or Radar data~\cite{wang_rdau-net_2022, zhang_3d_2023}. Traditional 3D CNNs are often constrained by high parameter counts, which lead to excessive memory consumption and slow training or inference~\cite{kopuklu_resource_2019}. To address these challenges, our 3D projections are processed with 2D CNNs which provide a lower-complexity alternative that significantly enhances processing speeds and reduces computational overhead~\cite{song_3dfeatures_2020, yoon_enhancement_2024}. Strategically placed channel- and spatial attention~\cite{woo_cbam_2018} helps in preserving the rich Doppler and Elevation features at a multi-scale level. With this work we want to advance the progress in Radar-only object detection under adverse weather conditions while motivating the use of  richer features from Radar FFT data for the training of deep learning models without being restricted by large dataset sizes. Therefore, the contributions of this paper are as follows:
\begin{itemize}
    \item Lightweight feature extraction network for a 3D projection of dense 4D Radar tensors using a CNN encoder-decoder structure in combination with spatial and channel attention.
    \item Decoupled detection heads adapted for heatmap-based center-point detection in Range-Azimuth coordinates combined with rotated 3D bounding box regression in cartesian world coordinates.
    \item Evaluation on various scenarios with multiple different road users and under challenging weather conditions on the large-scale K-Radar dataset.
\end{itemize}

\section{Background and Related Works}
\subsection{Radar Signal Processing}
We model the 4-dimensional Radar tensor $\mathcal{T}$ as superposition of $L$ targets, defined as $\mathcal{T}=\sum_{\ell=1}^L \mathcal{T}_{\ell}\in\mathbb{R}^{n_r\times n_a\times n_d\times n_e}$ with dimension of $n_r$, $n_a$, $n_d$ and $n_e$ comprising its Range, Azimuth, Doppler velocity and Elevation domain. The $\ell$th target $\mathcal{T}_{\ell}$ captures the unknown target parameters Range $R_\ell$, Azimuth $A_\ell$, Doppler velocity $D_{\ell}$ and Elevation $E_\ell$ encoded as low level Radar data at its $(R,A,D,E)$-th entry according to 
\begin{equation}
    [\mathcal{T}_\ell]=w_R(R-R_\ell)w_A(A-A_\ell)w_D(D-D_\ell)w_E(E-E_\ell)
\end{equation}
where the functions $w_R(R)$, $w_A(A)$, $w_D(D)$ and $w_E(E)$ describe the Radar sensor transfer function, known in advance and widely associated to Fourier-transformed window functions.

\subsection{Radar Object Detection}
Radar is frequently utilized as a complementary technology in sensor fusion approaches along with Lidar or Camera which provide more detailed resolution, essential for high object detection performance. Several works have investigated the fusion of Lidar and Radar point clouds for automotive object detection~\cite{huang_l4dr_2025, peng_mutualforce_2025, peng_moral_2025, wang_bi-lrfusion_2023}. However, while the point cloud format has proven successful for high resolution Lidar data, it is unsuitable for Radar data due to high sparsity and low angular accuracy which makes it challenging to extract meaningful patterns. Therefore, more recent and especially Radar-only approaches rely on FFT-processed RAD/RADE tensors~\cite{cheng_transrad_2025, dalbah_transradar_2024, fent_dpft_2024} or entirely on raw ADC data~\cite{giroux_t-fftradnet_2023}. Since these formats significantly increase the memory demands for each frame, the number of quality datasets that provide either RAD/RADE or ADC data is limited.

Notable datasets that explicitly provide raw ADC data include RadIal~\cite{rebut_raw_2022} and RaDICaL~\cite{lim_radical_2021} while K-Radar~\cite{paek_k-radar_2022}, UWCR~\cite{xm40-jx59-22}, and CARRADA~\cite{ouaknine_carrada_2021} offer processed RAD/RADE tensors. A comparison of these datasets reveals significant differences in terms of size, data format, and annotation quality. For instance, CARRADA is a relatively small dataset with 12666 frames (21.1 min) and stores RAD tensors in form of RA and RD maps along with 2D annotations. In contrast, UWCR only provides data from stationary ego perspective with object-center keypoint annotations derived from camera data. RADIal offers 25,000 synchronized frames, along with 2D bounding boxes in the image plane, including per-object distance and Doppler. RaDICaL provides data in various Radar configurations, along with ground truth point targets in the Radar Range-Azimuth heatmap obtained from the camera frame. To the best of our knowledge, K-Radar is the only available large scale dataset which provides RADE tensors (RAD + Elevation) as well as rotated 3D bounding boxes recorded under multiple different weather conditions. 

Building upon these datasets, several approaches have been proposed to improve Radar-only object detection, leveraging various architectures and techniques. For instance, TMVA-Net~\cite{ouaknine_multi-view_2021} and Ramp-CNN~\cite{gao_ramp-cnn_2021} utilize multi-view detection and 3D convolutional layers and are validated on the CARRADA and UWCR datasets. Other works, such as T-RODNet~\cite{jiang_t-rodnet_2023}, PeakConv~\cite{zhang_peakconv_2023}, and 3D UNet + ViT~\cite{zhang_3d_2023}, have explored different architectures on various datasets, including CARRADA, UWCR, and SCORP.

More recently, several studies have utilized the K-Radar dataset, which presents a more comprehensive and challenging environment for Radar-only object detection. CenterRadarNet~\cite{cheng_centerradarnet_2023} and RadaeOcc~\cite{ding_radarocc_2024} have demonstrated the effectiveness of 3D convolutional layers and data reduction techniques, respectively, on this dataset. Additionally, RTNH+~\cite{kong_rtnh_2023} and RTNH~\cite{paek_k-radar_2022} have proposed efficient encoding methods and preprocessing techniques to improve detection performance on K-Radar. The utilization of the K-Radar dataset has enabled researchers to develop more robust and accurate detection algorithms, which is essential for safe and autonomous navigation in adverse weather conditions.

\subsection{Model Prediction}
Radar-only detection approaches are evaluated on a variety of performance metrics depending on the type of dataset, data input, and tasks. Therefore, it is almost impossible to directly compare models across datasets. 
For instance, various efforts focused on semantic segmentation~\cite{ouaknine_multi-view_2021,gao_ramp-cnn_2021,jiang_t-rodnet_2023,zhang_peakconv_2023,zhang_3d_2023} where a class label is assigned to each point in the scene and the model prediction output is compared e.g., by their Intersection over Union (IoU) with the labels. Another task is to predict different types of bounding boxes in the scene and compare them to annotated ground truth boxes. Ultimately, there are 2D, 3D, rotated 2D, and rotated 3D boxes which bring different types of challenges. Approaches like~\cite{cheng_centerradarnet_2023,ding_radarocc_2024,paek_k-radar_2022,kong_rtnh_2023} work on rotated 3D bounding box regression where the model fits the size and orientation of each detected object as a bounding box with a semantic label.
Centerpoint~\cite{saini_centerpoint_2024} introduces the use of Transformers to improve center based bounding box prediction based on Radar RAD formats.

\section{Proposed Methodology}

\subsection{Overall Architecture}
\begin{figure}[htbp]
    \centering
    \includegraphics[width=\linewidth]{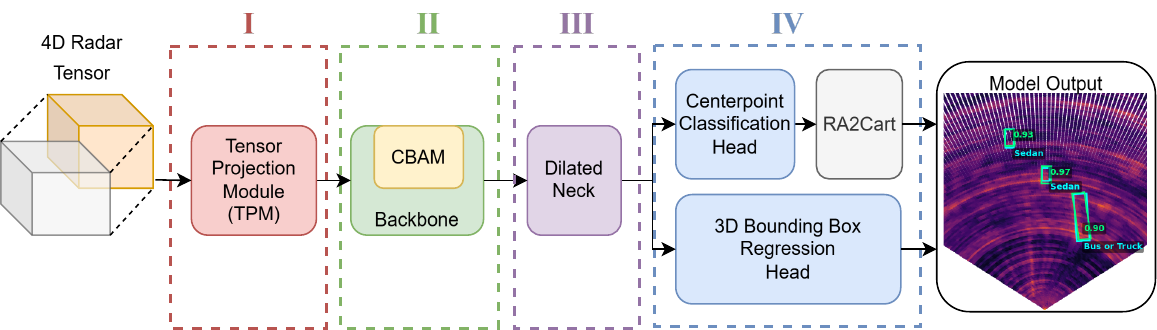}
    \caption{Proposed architecture: (I) Tensor Projection Module (TPM), (II) Encoder-Decoder Backbone, (III) Dilated Neck and (IV) Range-Azimuth based Centerpoint Detection Heads + Cartesian Transformation.}
    \label{fig:over_architecture}
\end{figure}

Our proposed architecture consists of four main building blocks shown in Fig.~\ref{fig:over_architecture}. First, the Tensor Projection Module (TPM) in Fig.~\ref{fig:tpm} creates the 3D projections from the full tensor. Since the TPM does not have learnable parameters, it can process the whole dataset at once and therefore reduce required memory during training by 91.9\% compared to loading a full tensor.  

\begin{figure*}[htbp]
    \centering
    \includegraphics[width=\linewidth]{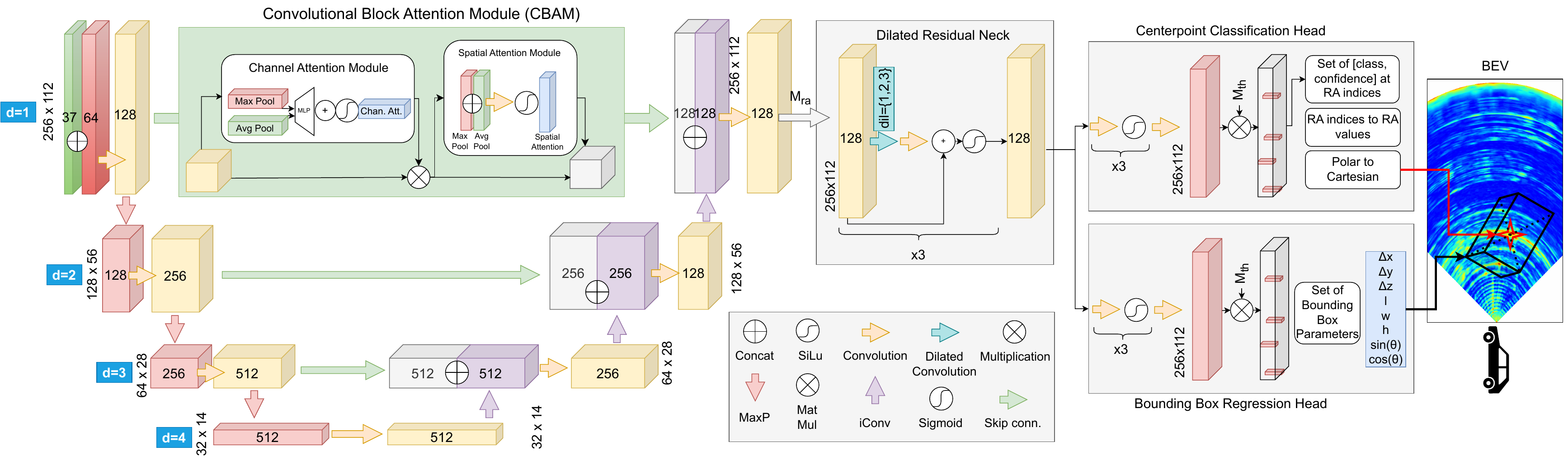}
    \caption{RADE-Net backbone with three down-sampling stages and CBAM supported skip connections.}
    \label{fig:backbone}
\end{figure*}

Second, the backbone in Fig.~\ref{fig:backbone} consists of a lightweight encoder-decoder structure which pairs high-level semantics (deep encoder features) with fine-grained localization (shallow features) via skip connections at different scales. In order to control the fusion of multi-scale Radar features between encoder and decoder, we adapted a spatial- and channel attention module from~\cite{woo_cbam_2018}. The final output of our backbone results in a dense Range-Azimuth feature map $M_{ra}\in \mathbb{R}^{n_r\times n_a\times 128}$ where Doppler and Elevation information is stored in the 128 feature channels for each RA bin.

Third, our dilated residual neck in Fig.~\ref{fig:backbone} is designed to increase the receptive field on $M_{ra}$ and take information from neighboring RA bins into account. This results in a processed feature map $M_{ra}'$.

Fourth, $M_{ra}'$ is forwarded to both decoupled heads independently, which is shown in Fig.~\ref{fig:backbone}. The center-point based classification head predicts a confidence map $M_{conf} \in [0,1]^{n_{cls}\times n_r\times n_a}$ for each class. Each Range-Azimuth bin in $M_{conf}$ is then transformed to obtain cartesian BEV center-points with respective confidence scores. The bounding box regression head predicts a set of 8 parameters $M_\varphi \in \mathbb{R}^{n_r\times n_a\times 8}$ which are used to fit a rotated 3D bounding box around each transformed BEV center-point with a sufficiently high confidence score. After the regression, overlapping boxes are removed by Non-Maxima Supression (NMS).

Finally, the reduced bounding boxes and corresponding class predictions are compared to annotated ground truth data to calculate the loss. We use Group Normalization \cite{wu_group_2018} with 32 groups to achieve normalization independent of batch size.

\subsection{Tensor Projection Module}
\begin{figure}[htbp]
    \centering
    \includegraphics[width=0.8\linewidth]{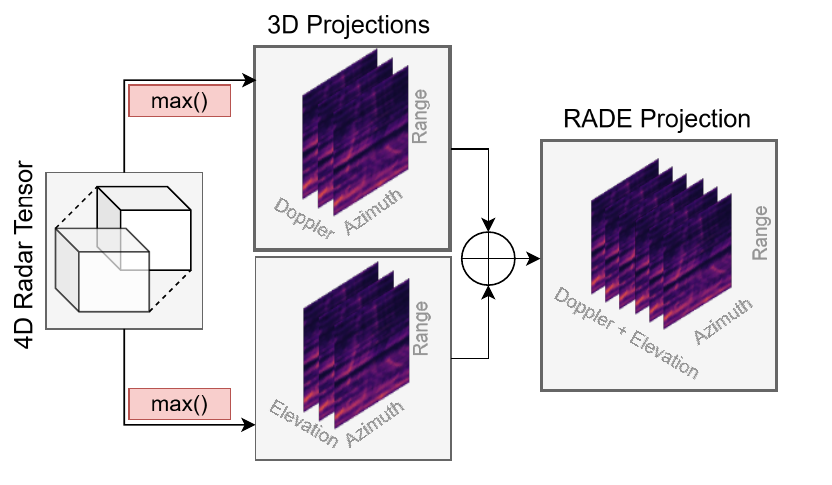}
    \caption{Tensor Projection Module (TPM): The full FFT-processed Radar tensor is split into Range-Azimuth-Doppler (RAD) and Range-Azimuth-Elevation (RAE) projections by calculating the maximum along each excluded dimension. The projections are subsequently concatenated along the third dimension.}
    \label{fig:tpm}
\end{figure}

We consider the FFT-processed Radar tensor $\mathcal{T}$ of a single frame in the K-Radar dataset. 
\begin{equation}
    \mathcal{T} \in \mathbb{R}^{n_r\times n_a\times n_d\times n_e}
\end{equation}
In order to preserve Doppler and Elevation information for each bin in the Range-Azimuth FoV, a combined 3D projection $\mathcal{P}^{ra'} \in \mathbb{R}^{(n_d+n_e)\times n_r\times n_a}$ is calculated as visualized in Fig.~\ref{fig:tpm}.
\begin{equation}
\label{eq:prade}
    \mathcal{P}^{ra} = \operatorname{Max}^E(\mathcal{T}) \oplus \operatorname{Max}^D(\mathcal{T})
\end{equation}
Since the downsampling stages in the backbone require both spatial dimensions to be divisible by the depth $d$, the final projection $\mathcal{P}^{ra} \in \mathbb{R}^{n_{de}\times n_r\times n_a}$ with $n_{a}=107$ is zero-padded to $n_{a}=112$. Both 3D projections result in a memory size of 21 MB per frame, compared to the the 260 MB per frame required by the full tensor, this results in a reduction of 91.9\%. Furthermore, the TPM processing time for a single frame is $1109 \pm 33ms$ (95\% confidence interval). However, this has not been optimized for real-time performance yet.

\subsection{Backbone}
The RADE-Net backbone consists of an encoder-decoder structure with a depth of 4 layers. Each layer of the encoder branch consists of a Max Pooling downsampling combined with a convolutional layer to compute the intermediate feature map
\begin{equation}
\label{eq:mid}
    M_{enc}^d \in \mathbb{R}^{(n_{de}\cdot d)\times (n_r/d)\times (n_a/d)}
\end{equation}
which is calculated as
\begin{equation}
  M_{enc}^{d}=\left\{
  \begin{array}{@{}ll@{}}
    \mathcal{P}^{ra}, & \ d=1 \\
    \operatorname{Conv}\big(\operatorname{MP}\big(M_{enc}^{(d-1)}\big)\big), & \ d = 2,3,4
  \end{array}\right.
\end{equation}
for all encoder-decoder skip connections. 

In order to control the flow of residual information in the skip connections, we adapted a Convolutional Block Attention Module (CBAM) from~\cite{woo_cbam_2018} and integrated it into our backbone as it was similarly approached by~\cite{wang_rdau-net_2022} for brain tumor segmentation on MRI data. The CBAM module sequentially applies a 1D channel attention map $A_c \in \mathbb{R}^{(n_{de}\cdot d)\times 1 \times 1}$ and a 2D spatial attention map $A_s \in \mathbb{R}^{1\times (n_r/d) \times (n_a/d)}$ on $M_{enc}^d$ at each depth $d$ of the backbone. This results in the residual feature map $M_r^d$ which is concatenated inside the decoder structure as shown in Fig.~\ref{fig:backbone}. Specifically, $M_r^d$ is computed as:
\begin{equation}
    {M_{enc}^d}' = A_c(M_{enc}^d) \otimes M_{enc}^d
\end{equation}
\begin{equation}
    M_r^d = A_s({M_{enc}^d}') \otimes {M_{enc}^d}'
\end{equation}
with $\otimes$ denoting the element-wise multiplication. Further, $A_c$ and $A_s$ are defined as
\begin{equation}
    A_c(M_{enc}^d) = \sigma(MLP(AP(M_{enc}^d)) + MLP(MP(M_{enc}^d)))
\end{equation}
\begin{equation}
    A_s({M_{enc}^d}') = \sigma(\operatorname{Conv}^{7\times7}(\operatorname{AP}({M_{enc}^d}') \oplus \operatorname{MP}({M_{enc}^d}')))
\end{equation}
where $\operatorname{AP}$ denotes Average Pooling, $\operatorname{MP}$ Max Pooling, $\sigma$ the sigmoid function and $\oplus$ concatenation operation. A single two-layer feed-forward MLP is shared and applied to the outputs of both the MP and AP branches. For a branch output $B_o \in \mathbb{R}^{C  \times 1 \times 1}$, we flatten it to $x \in \mathbb{R}^n$ with $n = C \cdot 1 \cdot 1$ and compute
\begin{equation}
    MLP(x) = L_2(ReLU(L_1(x)))
\end{equation}
where $L_1: \mathbb{R}^n \rightarrow \mathbb{R}^d$ and $L_2: \mathbb{R}^d \rightarrow \mathbb{R}^n$, with $d = \lfloor \frac{n}{16} \rfloor$. We then shape the output back and apply the channel weights.

For the decoder branch, each intermediate feature map $M_{dec}^d$ is calculated by an inverse convolution for spatial up-sampling followed by concatenation with the residual feature map $M_r^d$ from the CBAM and a subsequent convolution for "channel compression".
\begin{equation}
\label{eq:decoder}
  M_{dec}^{d}=\left\{
  \begin{array}{@{}ll@{}}
    M_{enc}^{d}, & \ d=4 \\
    \operatorname{Conv}\big(M_r^d\oplus \operatorname{iConv}\big(M_{dec}^{(d+1)}\big), & \ d = 1,2,3
  \end{array}\right.
\end{equation}
Finally, the dense Range-Azimuth feature map $M_{ra}$ is given by $M_{dec}^1$ and further passed to the neck.

\subsection{Dilated Residual Neck}
We implemented a residual neck with dilated convolutions~\cite{yu_multi-scale_2016} to increase the receptive field on the dense Range-Azimuth feature map $M_{ra}$. The neck is built from three residual blocks consisting of dilated convolutions $\operatorname{Conv}_{dil=k}$ with dilation $k=(1,2,3)$, followed by a regular convolution and SiLU activation function. Therefore, each block is defined as
\begin{equation}
    \operatorname{Res}^k(M)=\operatorname{SiLU(}M + (\operatorname{Conv}_{dil=1}(\operatorname{Conv}_{dil=k}(M)))
\end{equation}
and is combined into the dilated residual neck.
\begin{equation}                
    M_{ra}'=\operatorname{Res}^3(\operatorname{Res}^2(\operatorname{Res}^1(M_{ra})))
\end{equation}
\subsection{Detection Heads}
Our proposed decoupled heads are inspired by CenterNet~\cite{zhou_objects_2019} and further adapted for detection in Range-Azimuth coordinates and rotated 3D bounding box regression in cartesian coordinates. Therefore, the detection and regression head both take $M_{ra}'$ as an input. For the classification head, first, a center-point confidence map $M_{conf} \in [0,1]^{n_{cls}\times n_r\times n_a}$ is predicted for each class $c_i$ by applying three convolutions on $M_{ra}'$, each followed by a Sigmoid activation function. Further, a threshold mask
\begin{equation}
    M_{th}(c,r, a)=\left\{
  \begin{array}{@{}ll@{}}
    1, & \text{if}\ M_{ra}'(c,r,a) \geq \tau_{cls}\\
    0, & \text{else}
  \end{array}\right.
\end{equation}
is applied on all non-background classes to get the valid Range-Azimuth center-point map $M_{cp}$.
\begin{equation}
    M_{cp} = M_{conf}\otimes M_{th}
\end{equation}
The non-zero entries in $M_{cp}$ are then converted to their respective Range and Azimuth value and further transformed into a list of center-points $\mathbb{P}=\{p_1, ..., p_n\}$ in cartesian coordinates where $p_i=(x_p, y_p)$. 

In order to fit the bounding box around each center-point $(x_{cp},y_{cp},z_{cp})$, the 3D bounding box regression head applies three convolutional layers to $M_{ra}'$, producing a parameter map $M_\varphi \in \mathbb{R}^{n_r\times n_a\times 8}$. Each vector in $M_\varphi$ encodes eight parameters that define the 3D bounding box. Specifically, for every detected center-point $\mathbb{P} = (x_{cp}, y_{cp}, z_{cp})$, the network predicts three local offset parameters $(\Delta x, \Delta y, \Delta z)$, which are regressed directly by the bounding box regression head. These offsets achieve sub-bin resolution of the discretized center-point locations. The corrected center-point coordinates are calculated as
\begin{equation}
    \mathbb{P}' = (x_{cp}', y_{cp}', z_{cp}') = (x_{cp}+\Delta x, y_{cp}+\Delta y, z_0 +\Delta z)
\end{equation}
where $z_0$ equals the Radar sensor height offset above ground. Further, the parameters $(l, w, h)$ describe the length, width and height of the bounding box and $(\sin{\theta}, \cos{\theta})$ are used as a numerical stable representation of the orientation $\theta$. 

After performing the regression for each center-point in $M_{cp}$ we arrive at a full set of bounding boxes $\mathbb{B} = \{b_1, ..., b_m\}$. In order to avoid multiple bounding boxes for a single object, Non-maxima suppression (NMS) with an IoU threshold of 0.3 is performed in the inference stage to compute a set of reduced bounding boxes $\mathbb{B}_{nms}$. Furthermore, other than the convolutional layers, both heads do no contain additional trainable parameters.

\subsection{Loss Function}
Our model is trained with a combination of three loss components. A focal loss optimizes the center-point prediction, and a composite regression loss, combining the Gaussian-Wasserstein Distance (GWD) and a smooth L1 term, optimizes the predicted bounding boxes. All components are normalized by their mini-batch mean to achieve uniform gradient updates. To emphasize accurate center-point localization, we assign a weight of two to the focal loss.
\begin{equation}
    \mathcal{L}_{all}=2\cdot\mathcal{L}_{foc}+\mathcal{L}_{gwd}+\mathcal{L}_{L1}
\end{equation}
\subsubsection{Focal Loss}
We design a modified cross-entropy (CE) loss which operates on continuous probability maps rather than binary classifications. The implementation generates Gaussian distributions centered at the ground truth bounding box locations, using $\sigma = 3$, to construct a multiclass focal ground truth map $M_{foc,gt}$ with the same shape as $M_{conf}$. The CE loss is then calculated between $M_{conf}$ and $M_{foc,gt}$, with the focal loss parameters set to $\alpha = 2$ and $\gamma = 4$. Furthermore, we apply a custom implementation of $\text{reduction} = \text{'mean'}$ across all dimensions.
\subsubsection{Gaussian-Wasserstein Distance Loss}
Most common approaches rely here on a 3D Intersection over Union (IoU) based loss, which requires computational expensive algorithms for rotated 3D bounding boxes that may lose differentiability. GWD loss~\cite{yang_rethinking_2022} is a computationally efficient, differentiable regression loss function that treats bounding boxes as Gaussian distributions, where the mean represents the center coordinates and the covariance matrix encodes the box dimensions and orientation, and computes the Wasserstein distance between predicted and ground truth distributions. When computing the matrix square root, we use eigenvalue decomposition. Following~\cite{yang_rethinking_2022}, we apply a non-linear transformation, specifically opting for the square root function with an empirically selected parameter $\tau = 1.65$.
\subsubsection{Smooth L1 Loss}
The loss is applied element-wise across all 8 bounding box parameters for each box in the set of $\mathbb{B}_{nms}$ with a matching ground truth box in $\mathbb{B}_{gt}$. We use the default setting of $\beta = 1$ and $\text{reduction} = \text{'mean'}$. The final L1 loss is then averaged over the matching set with zero loss for empty cases.

Because the Gaussian density is concentrated near the center, the GWD term tends to emphasize agreement near the central region. Once the centers are aligned, the gradients with respect to the covariance matrix tend to be relatively small. In practice, we observed that this bias leads the model to conservative predictions that did not cover ground truth boxes to its full extent (i.e., predicted bounding boxes were consistently smaller and under-covered the area). To counteract this, we introduced the Smooth L1 penalty to form a composite loss for our bounding box regression.

\section{Experiments}
\subsection{Dataset}
Our model is trained and evaluated on the large scale K-Radar dataset~\cite{paek_k-radar_2022} which contains 35K frames of 4D Radar tensors, annotated 3D bounding box labels of various different road users, additional 3D Lidar point clouds and camera images. The dataset covers different driving scenarios (urban, suburban roads, alleyways, and highways) and weather scenarios (overcast, fog, rain, and snow). Each frame is initially represented by a FFT-processed Radar tensor $ \mathcal{T} \in \mathbb{R}^{256 \times 107 \times 64 \times 37}$. Considering all known Radar transfer functions, this results in a Radar FoV with a maximum range of 118m, an azimuth FoV of 107° and an elevation FoV of 37°. All additional Radar parameters can be found in~\cite{paek_k-radar_2022}. 

\begin{table*}[htbp]
    \centering
    \caption{Sedan detection comparison under various weather conditions}
    \begin{tabular}{c|c|c|cccccccc}
    \hline
        Modality & Metric & Networks & Total & Normal & Overcast & Fog & Rain & Sleet & Light Snow & Heavy Snow\\ \hline \hline
        \multirow{3}{*}{Lidar} & \multirow{3}{*}{$AP_{3D}$} & Voxel-RCNN \cite{deng_voxel_2021} & 46.4 & 81.8 & 69.6 & 48.8 & 47.1 & 46.9 & 54.8 & 37.2\\
        &  & CasA \cite{wu_casa_2022} & 50.9 & 82.2 & 65.6 & 44.4 & 53.7 & 49.9 & 62.7 & 36.9 \\
        &  & TED-S \cite{wu_transformation-equivariant_2022} & 51.0 & 74.3 & 68.8 & 45.7 & 53.6 & 44.8 & 63.4 & 36.7 \\
        \hline
        \multirow{3}{*}{ \begin{tabular}{@{}c@{}} Camera \\ +Lidar \end{tabular} } & \multirow{3}{*}{$AP_{3D}$} & VPFNet \cite{zhu_vpfnet_2021} & 52.2 & 81.2 & 76.3 & 46.3 & 53.7 & 44.9 & 63.1 & 36.9 \\
        &  & TED-M \cite{wu_transformation-equivariant_2022} & 52.3 & 77.2 & 69.7 & 47.4 & 54.3 & 45.2 & 64.3 & 36.3 \\
        &  & MixedFusion \cite{zhang_mixedfusion_2025} & 55.1 & \textbf{\phantom{*}84.5*} & \textbf{\phantom{*}76.6*} & 53.3 & 55.3 & 49.6 & \textbf{\phantom{*}68.7*} & 44.9 \\
        \hline
        Camera & \multirow{2}{*}{$AP_{3D}$} & EchoFusion \cite{liu_echoes_nodate} & 47.4 & 51.5 & 65.4 & 55.0 & 43.2 & 14.2 & 53.4 & 40.2 \\
        +Radar& & DPFT \cite{fent_dpft_2024} & 56.1 & 55.7 & 59.4 & 63.1 & 49.0 & 51.6 & 50.5 & 50.5 \\
        \hline
        \multirow{6}{*}{Radar} & \multirow{3}{*}{$AP_{3D}$} & RTNH \cite{paek_k-radar_2022} & 47.4 & 49.9 & 56.7 & 52.8 & 42.0 & 41.5 & 50.6 & 44.5\\
        & & RTNH+ \cite{kong_rtnh_2023} & 57.6 & - & - & - & - & - & - & -\\
        & & \textbf{RADE-Net (Ours)} & \textbf{\phantom{*}64.1*} & 60.7 & 72.0 & \textbf{\phantom{*}85.4*} & \textbf{\phantom{*}55.4*} & \textbf{\phantom{*}63.6*} & 68.6 & \textbf{\phantom{*}67.6*}\\ \cline{2-11}
        & \multirow{3}{*}{$AP_{BEV}$} & RTNH \cite{paek_k-radar_2022} & 58.4 & 58.5 & 64.2 & 76.2 & 58.4 & 60.3 & 57.6 & 56.6\\
        & & RTNH+ \cite{kong_rtnh_2023} & 65.7 & - & - & - & - & - & - & -\\
        & & \textbf{RADE-Net (Ours)} & \textbf{\phantom{**}68.7**} & \textbf{\phantom{**}65.2**} & \textbf{\phantom{**}74.6**} & \textbf{\phantom{**}91.3**} & \textbf{\phantom{**}63.1**} & \textbf{\phantom{**}67.9**} & \textbf{\phantom{**}74.1**} & \textbf{\phantom{**}68.7**}\\ \hline
    \end{tabular}
    \\
    \small *best performance in $AP_{3D}$ **best performance in $AP_{BEV}$
    \label{tab:performance_comparison}
\end{table*}

\subsection{Experiment Setup}
\subsubsection{Training}
The training is performed on a single NVIDIA L30, optimized using AdamW with an initial learning rate of 0.001 and weight decay of 0.01. To ensure stable convergence, a cosine annealing learning rate scheduler was applied with a minimum learning rate of 0.0001. The model is trained with the official K-Radar train-test split in two configurations: First, only on Sedan (Car) objects for 10 epochs to be consistent with the K-Radar baseline models~\cite{paek_k-radar_2022, kong_rtnh_2023} and other approaches listed in Table~\ref{tab:performance_comparison}. Second, on all object classes for general evaluation under different road and weather conditions.

\subsubsection{Inference}
To assess real-time performance, we report an inference latency measured on batches of 10 frames processed on a single NVIDIA L40S GPU, using 561 samples. The measured latencies are as follows: backbone, $49ms$; neck, $31ms$; and head, $18ms$.

\subsubsection{Evaluation Metrics}
For performance comparison, we implemented the popular mean Average Precision (mAP) which combines precision and recall in a single metric across multiple object classes. For single class evaluation in Table~\ref{tab:performance_comparison} we show just the Average Precision (AP). We follow the K-Radar guideline and consider a Region-of-interest (ROI) for detection of $x\in[0,72m]$, $y\in [-6.4m, 6.4m]$ and $z\in [-2m, 6m]$.

\subsection{Performance Comparison}
Table~\ref{tab:performance_comparison}  presents our model’s performance across all weather conditions, comparing 3D AP against various implementations using different modalities and combinations such as Lidar, Camera + Lidar , Camera + Radar and Radar. Since the official K-Radar baseline is additionally evaluated on BEV AP, this comparison is included in Table~\ref{tab:performance_comparison}. RTNH+~\cite{kong_rtnh_2023} has not been evaluated on different weather conditions and therefore we compare only the total AP.

%
%
%
%
%
%
%
%
%
%


\begin{table*}[htbp]
    \centering
    \caption{Evaluation of our RADE-Net with multiple road users and under various weather conditions}
    \begin{tabular}{c|cc|cc|cc|cc|cc|cc|cc}
    \hline
        \multirow{3}{*}{Object class} & \multicolumn{14}{|c}{Average Precision (AP)}\\ \cline{2-15}
         & \multicolumn{2}{|c|}{Total} & \multicolumn{2}{|c|}{Normal} & \multicolumn{2}{|c|}{Fog} & \multicolumn{2}{|c|}{Rain} & \multicolumn{2}{|c|}{Sleet} & \multicolumn{2}{|c|}{Light Snow} & \multicolumn{2}{|c}{Heavy Snow}\\
        &3D & BEV & 3D & BEV & 3D & BEV
        & 3D & BEV & 3D & BEV & 3D & BEV & 3D & BEV \\ \hline \hline
        Sedan           & 56.75 & 63.21 & 55.10 & 61.29 & 83.09 & 89.13 & 49.11 & 58.69 & 39.96 & 44.45 & 62.10 & 72.22 & 61.82 & 63.33 \\ 
        Bus or Truck    & 40.99 & 48.29 & 37.25 & 40.77 & -     & -     & 1.05  & 1.05  & 14.12 & 17.99 & 68.96 & 72.56 & 65.94 & 85.94 \\ 
        Pedestrian      & 5.40  & 6.79  & 0.99  & 0.99 & 15.43 & 19.94 & 0.99  & 0.99  & 1.40  & 1.60  & -     & -     & -     & - \\ 
        Bicycle         & 1.49  & 1.49  & 1.49  & 1.49     & -     & -     & -     & -     & -     & -     & -     & -     & -     & - \\ 
        Total (mAP)     & 26.16 & 29.95 & 23.71 & 26.13 & 49.26 & 54.53 & 17.05 & 20.24 & 18.52 & 21.53 & 65.53 & 72.39 & 63.88 & 74.63 \\ \hline
    \end{tabular}
    \label{tab:performance_multiobject}
\end{table*}

\subsection{Results and Discussion}
\begin{figure*}[htbp]
    \centering
    \includegraphics[width=\linewidth]{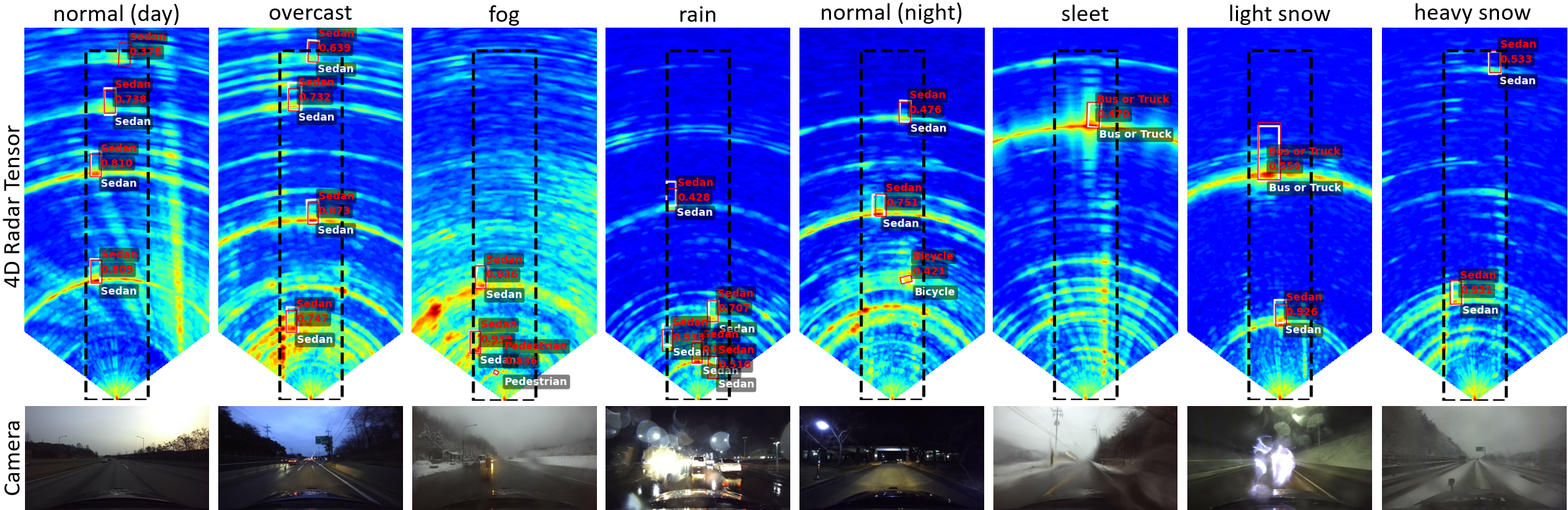}
    \caption{Model performance visualization under different weather conditions. The ground truth bounding boxes are shown in white along with class label and the model predictions are shown in red along with class label and confidence score. The ROI where detections are considered is marked with a dashed black rectangle.}
    \label{fig:model_vis}
\end{figure*}

In Fig.~\ref{fig:model_vis} we show qualitative results of RADE-Net in different weather conditions along with camera view. The results demonstrate a robust detection regardless of noise induced by adverse weather or angular uncertainty. According to the results in Table~\ref{tab:performance_comparison}, our model significantly outperforms the Radar-only baseline as well as Camera-Radar fusion approaches in all weather conditions and the RTNH+ model by 5.4\% in total AP.

\begin{table*}[htbp]
    \centering
    \caption{Ablation Study}
    \begin{tabular}{cccccc|c|c|c|c}
    \hline
    Feature & Input & Dilated & Expanded & \multirow{2}{*}{CBAM} & \multirow{2}{*}{Base} & \multicolumn{2}{|c|}{IoU=0.3} & \multicolumn{2}{|c}{IoU=0.5} \\ \cline{7-10}
    Expansion & Stem & Neck & Heads & & & $AP_{3D}$ & $AP_{BEV}$ & $AP_{3D}$ & $AP_{BEV}$\\ \hline \hline 
    & & & & & $\checkmark$& 57.96 & 63.98 & 32.85 & 56.18\\
     & & & & $\checkmark$ & $\checkmark$ & 55.13 & 62.88 & 27.65 & 50.99 \\
   & & & $\checkmark$ & $\checkmark$ & $\checkmark$ & 58.35 & 64.70 & 32.73 & 55.98 \\
   & & $\checkmark$ & $\checkmark$ & & $\checkmark$ & 59.81 & 66.32 & 35.31 & 57.01 \\
    & & $\checkmark$ & $\checkmark$ & $\checkmark$ & $\checkmark$ & \textbf{64.06} & 68.71 & \textbf{39.02} & \textbf{61.64}\\
    & $\checkmark$ & $\checkmark$ & $\checkmark$ & $\checkmark$ & $\checkmark$ & 63.47 & \textbf{69.05} & 37.09 & 60.82 \\
    $\checkmark$ & & $\checkmark$ & $\checkmark$ & $\checkmark$ & $\checkmark$ & 63.12 & 67.92 & 38.04 & 60.22 \\
    \hline
    \end{tabular}
    \label{tab:ablation_study}
\end{table*}

Due to their high accuracy and point-cloud density, Lidar and Camera-Lidar models perform high on normal weather conditions and therefore surpass all Radar-based approaches. However, conditions like fog, rain, sleet and snow block Lidar signals and induce significant noise to their measurement which causes the performance to decrease drastically. Meanwhile it can be observed that the performance of Radar-only or Camera-Radar models is less affected by adverse weather conditions. Our model surpasses Lidar and Camera-Lidar models in almost all adverse weather condition categories.

Table~\ref{tab:performance_multiobject}  shows our evaluation on four different road users. Once other classes are introduced, we observe a lower performance for the "Sedan" class detection. Here, the most challenging task for the model is to differentiate between the classes "Sedan" and "Bus or Truck", since both cause a similar reflection characteristic to the Radar sensor. The data indicates a low detection performance for pedestrians and bicycles which is primarily due to their low Radar cross section, occlusion by angular artifacts of larger objects, and general under-representation in the dataset~\cite{paek_k-radar_2022}.

Furthermore, the results indicate that all Radar and Radar-Camera models perform notably better in adverse weather scenarios (e.g., fog, overcast, snow, ...) compared to normal scenarios. This counter-intuitive finding can be attributed to the distribution of labeled bounding boxes across weather conditions and detection distances, which is thoroughly  analyzed and discussed in~\cite{huang_l4dr_2025}.

\subsection{Ablation Study}
We trained different model configurations to show the effect of single building blocks in the architecture. The configurations are set for sedan-only classification and listed in Table~\ref{tab:ablation_study} with BEV and 3D AP for IoU thresholds of 0.3 and 0.5 using the same training parameters and ROI as for the performance comparison in Table~\ref{tab:performance_comparison}. Specifically, we ablate (i) CBAM, i.e., the spatial and channel attention module shown in Fig.~\ref{fig:backbone}; (ii) the expanded heads, defined as the three successive convolutional layers with activation functions applied to $M_{ra}'$ by each head; and (iii) the dilated neck, comprising the dilated convolutions and the residual connection shown in Fig.~\ref{fig:backbone}. 

In addition, we evaluate two implementation modifications: an input stem and feature expansion. For the input stem, the 3D projections in Fig.~\ref{fig:tpm} are processed separately for Doppler and Elevation using a convolutional layer prior to concatenation. For feature expansion, the convolution in the final decoder stage $M_{dec}^1$ in~\eqref{eq:decoder} is increased to 256 feature channels instead of 128, with corresponding dimensional adaptations in the neck and heads.

Table~\ref{tab:ablation_study} shows that the spatial and channel attention modules within the skip connections result in a significant performance increase, but only when the neck and heads are complex enough. Since the feature expansion leads to a decrease in AP, we argue that a feature dimension of 128 is sufficient for our model tasks.

\section{Conclusion}
We successfully developed a method to preserve inherent relations between spatial Range-Azimuth bins and Doppler as well as Elevation features in Radar FFT-processed tensors while limiting the required data size for training. Our model backbone combines spatial and channel attention with a 2D CNN encoder-decoder structure, effectively processing 3D Radar projections while reducing the high model complexity associated with 3D CNNs. The detection head is adapted to find center-points in the Range-Azimuth domain which preserves the original polar coordinates of the Radar data tensor and then regress oriented 3D bounding boxes in the transformed cartesian coordinate frame. Our model is evaluated under various weather conditions and compared to several single- and multi-modal approaches. Notably, we outperform the K-Radar baseline by 16.7\% and their most recent Radar-only approach by 6.5\% in overall AP. Moreover, our method outperforms LiDAR-based approaches across all adverse weather conditions, with the largest gain of 32.1\% in foggy scenarios. Future research directions may include a focus on improving Radar-only detection performance for smaller objects in the dataset such as pedestrians and cyclists.

\section*{Acknowledgment}

This work is supported by Infineon Technologies Austria AG and the European Union through the Horizon Europe programme (Grant Agreement project 101092834). Funded by the European Union.

\bibliographystyle{ieeetr}
\bibliography{bibliography}

\end{document}